# NOISE ANALYSIS FOR LENSLESS COMPRESSIVE IMAGING

Hong Jiang, Gang Huang and Paul Wilford

Bell Labs, Alcatel-Lucent, Murray Hill, NJ 07974

*Abstract*— We analyze the signal to noise ratio (SNR) in a recently proposed lensless compressive imaging architecture. The architecture consists of a sensor of a single detector element and an aperture assembly of an array of aperture elements, each of which has a programmable transmittance. This lensless compressive imaging architecture can be used in conjunction with compressive sensing to capture images in a compressed form of compressive measurements. In this paper, we perform noise analysis of this lensless compressive imaging architecture and compare it with pinhole aperture imaging and lens aperture imaging. We will show that the SNR in the lensless compressive imaging is independent of the image resolution, while that in either pinhole aperture imaging or lens aperture imaging decreases as the image resolution increases. Consequently, the SNR in the lensless compressive imaging can be much higher if the image resolution is large enough.

## I. INTRODUCTION

LENSLESS compressive imaging (LCI) [1] is an effective architecture to acquire images using the compressive sensing technique [2][3]. It consists of a sensor of a single detecting element and an aperture assembly of programmable aperture elements, but no lens is used, as illustrated in Figure 1. The transmittance of each aperture element is individually programmable. The sensor can be used to acquire compressive measurements which, in turn, can be used to reconstruct an image of the scene. By using compressive sensing, an image can be reconstructed using far fewer measurements than the number of pixels in the image, and therefore, an image is already compressed when it is acquired in the form of compressive measurements. This architecture is distinctive in that the images acquired are not formed by any physical mechanism, such as a lens [4] or a pinhole [5]-[7]. This results in the feature that there are no aberrations introduced by a lens, such as a scene being out of focus. Furthermore, the same architecture can be used for acquiring multimodal signals such as infrared, Terahertz [8] and millimeter wave images [9]. This architecture has application in surveillance [10].

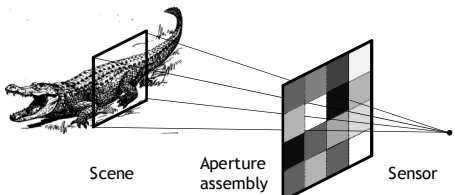

**Figure 1.** Lensless compressive imaging (LCI) architecture

Furthermore, multi-view imaging can be easily accommodated by placing multiple sensors in conjunction with one aperture assembly [11]. Compressive measurements from the multiple sensors can be used in a variety of ways. For example, they may be used to reconstruct multi-view images such as 3D images. Measurements from multiple sensors may also be used in a joint reconstruction to form a single image, for the purposes of 1) reducing number of measurements taken by each sensor, 2) enhancing quality of the reconstructed images by concatenating measurements from all sensors, and 3) increasing the resolution of the image beyond the resolution of the aperture assembly; see [11] for details.

*A. Difference from other imaging architectures*

The LCI architecture of [1], as shown in Figure 1, is different from most other known imaging architectures, such as pinhole aperture imaging (PAI) [5], coded aperture imaging [6][7] or single-pixel imaging with lens aperture [4]. One fundamental difference is that in the LCI of [1], no physical images are formed, while in all other referenced imaging architectures, a physical image is formed either by a lens [4] or by a pinhole [5]-[7]. This fundamental difference underscores the different functionality of the apertures. Since no physical image is formed, the functionality of the aperture assembly in Figure 1 is simply to transmit or to block light, and there is little other constraints placed on the aperture elements; for example, there is no constraint on the size of the aperture elements, and the size of the elements does not affect the quality of the image. On the other hand, the functionality of apertures for other imaging architectures [4]-[7] is to form an image, and therefore, the apertures must meet certain restrictive requirements to provide an image of desired quality; for example, there are constraints on the curvature of a lens or the size of a pinhole in order to form a sharp image. Even the coded aperture [6][7] is different from the aperture assembly of Figure 1 because each element in the coded aperture acts as a pinhole to form one view of the image, and therefore, its size affects the quality of the reconstructed image, while the image quality in the LCI of Figure 1 is independent of the size of its aperture elements.

Furthermore, in pinhole aperture imaging or coded aperture imaging [5]-[7], a large array of sensors must be used, one sensor for each pixel in the image, while only one sensor is used in LCI of Figure 1.

*B. Main contribution of this paper*

Since the LCI architecture of Figure 1 does not use a lens, does it suffer from the poor signal to noise ratio (SNR), and how does its SNR compare to that of a pinhole aperture imaging or to that of a traditional digital camera with a lens (lens aperture imaging)?

The goal of this paper is to answer this question. We will perform a noise analysis and compare the lensless compressive imaging with the pinhole aperture imaging and the lens aperture imaging, e.g., a traditional digital imaging with a lens.

The LCI architecture allows an image to be acquired directly as compressed data, in the form of compressive measurements. There are two types of noises in the final reconstructed image: measurement noise and compression noise. The measurement noise is defined as the noise present in the process of acquiring data from the imaging device, such as shot noise, thermal noise and quantization noise in the acquired data. The compression noise is defined as noise due to compression, i.e., the error introduced in the reconstruction because not all independent compressive measurements are used, even if the measurements themselves are acquired precisely, free of any measurement noise.

Similarly, in addition to the measurement noise, the pinhole aperture imaging or lens aperture imaging may also suffer from the compression noise, which is the error introduced when an image is compressed to, e.g., JPEG format.

In this paper, we only analyze the measurement noise. The analysis of compression noise, i.e., errors due to use of partial measurements in reconstruction in the lensless compressive imaging, can be found in general compressive sensing literature, e.g., [2][3]. In neglecting the compression noise, we assume all independent measurements are used in reconstruction and analyze SNR of the reconstructed image due to the measurement noise. Two types of noises are included in the measurement noise. The first type is the shot noise, due to statistical quantum fluctuations, which is modeled by a Poisson distribution. The second type is additive noise which includes thermal noise, quantization noise etc. and is modeled by a random variable of zero mean and certain variance. There is no assumption on the type of distribution for additive noise.

Our analysis will show that the SNR of the lensless compressive imaging can be much better than that of pinhole aperture imaging, and furthermore, it can even be better than that of the lens aperture imaging. More specifically, how the SNR of the LCI compares with that of, say, the traditional digital imaging with a lens (lens aperture imaging), depends on, among other parameters, the resolution of the image. Our analysis will show that, when other parameters, such as noise characteristics of sensors, the size of lens aperture etc, are fixed, the LCI architecture of Figure 1 will eventually have a better SNR than the lens aperture imaging if the resolution, i.e., the number of pixels, of the image is large enough. We will provide insights and explanations on why this result makes intuitive sense.

*C. Acronym*

The following acronyms will be used extensively throughout the rest of the paper.

**LAI** – Lens aperture imaging. This refers to the imaging architecture in which a lens aperture is used to form an image which is pixelized by an array of sensors. An example is a legacy digital camera with a lens.

**LCI** – Lensless compressive imaging. This refers to the imaging architecture shown in Figure 1, which was first proposed in [1].

**MN** – Measurement noise. This refers to the noise in the data obtained from sensor or sensors, including such noises as shot noise, thermal noise and quantization noise. This is in contrast to compression noise which is the error introduced in the process of compressing images. We consider measurement noise only in this paper.

**PAI** – Pinhole aperture imaging. This refers to the imaging architecture in which a pinhole aperture is used to form an image which is pixelized by an array of sensors.

**SNR** – Signal to noise ratio.

*D. Organization of the paper*

The paper is organized as follows. In Section II, we review the LCI architecture. Then, the SNR of the architecture is analyzed in Section III under the assumption of measurement noise only. In Section IV, comparison is made with PAI and LAI. Further discussions are provided in Section V to facilitate better understanding of the LCI architecture. Simulation results are reported in Section VI, which is followed by the Conclusion.

II. LENSLESS COMPRESIVE IMAGING

In this section, we provide a review of the LCI architecture of [1].

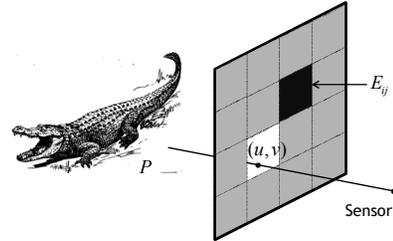

**Figure 2.** Illustration for the definition of virtual images in lensless compressive imaging.

As shown in Figure 2, for each point $(u,v)$ on the aperture assembly, there is a ray starting from a point, $P$, on the scene, passing through the point $(u,v)$, and ending at the sensor. Denote by $r(u,v;t)$ the intensity of the unique ray associated with point $(u,v)$ on the aperture assembly at time $t$. An image $x(u,v)$ of the scene detected by the sensor is defined as the integration of the ray in a time interval $\Delta t$:

$$x(u,v) = \int_0^{\Delta t} r(u,v;t)dt . \quad (1)$$

The image in (1) is called a virtual image because there is not an actual image formed by any physical mechanism. The virtual image can be pixelized by the aperture assembly. Let an aperture element be denoted by $E_{ij}$. Then the pixel value at the pixel $(i,j)$ is given by

$$x(i,j) = \iint_{E_{ij}} x(u,v)dudv . \quad (2)$$

The image $x(i,j)$ has as many pixels as the number of elements in the aperture assembly. When the aperture assembly is programmed to implement a compressive sensing matrix, the transmittance of each aperture element is set to equal the value of the corresponding entry in the sensing matrix. Let the sensing matrix $A$ be a random matrix whose entries, $a_{nm}$, are random numbers between 0 and 1, and let $T^m(u,v)$ be the transmittance programmed according to row $m$ of $A$. Then the compressive measurements are given by

$$y_m = \iint T^m(u,v)x(u,v)dudv = \sum_{i,j} a_{m,q(i,j)} x(i,j), \text{ or} \quad (3)$$
$$y = Ax,$$

where $q$ is mapping from a 2D array to a 1D vector, and $y$ is the measurement vector, $A$ is the sensing matrix and $x$ is the vector representation of the pixelized image $x(i,j)$.

In the rest of the paper, we will treat the image $x$ either as a 2D array or as a 1D vector, interchangeably. In particular, the notation $x(i,j)$ or $x_{ij}$ refers to an element in the 2D array, and $x(i)$ or $x_i$ refers to an element in 1D vector.

### III. SIGNAL TO NOISE RATIO

In this section, we perform SNR analysis for the LCI. As noted in the Introduction, we will only consider the measurement noise, the noise that is introduced into the measurements while they are acquired. The compression noise, the error due to reconstruction by using only portion of measurements, is not considered in this paper, and its analysis has been performed extensively in the compressive sensing literature such as [2][3]. Therefore, in this paper, we will assume that the sensing matrix is a square matrix, i.e., there are as many measurements as the number of pixels in an image, and all measurements are used in the reconstruction of the image.

*A. Sensing matrix and measurements*

To make measurements, the transmittance of the aperture elements in Figure 1 is programmed according to a sensing matrix. In theory, a sensing matrix needs to have the restricted isometry property (RIP) [3][12], but in practice, a randomly permutated, modified, Hadamard matrix is usually used with satisfactory result [1][10][11][13]. Since the random permutation does not change the noise analysis, we will use a modified Hadamard matrix as the sensing matrix. We need to modify the Hadamard matrix, which has entries of -1s and 1s, because the transmittance is not defined for -1.

Let $N$ be the total number of aperture elements, which is also the total number of pixels in the reconstructed image. Let $A = [a_{ij}]$ be an $N \times N$ matrix created by Hadamard matrix $H = [h_{ij}]$ with $-1$s replaced by $0$s, i.e., the entries of $A$ are given by

$$a_{ij} = \begin{cases} 1, & h_{ij} = 1 \\ 0, & h_{ij} = -1 \end{cases}. \quad (4)$$

Let $x$ be the vector of length $N$ whose component $x_i$ is the light intensity from pixel $i$ of the object plane. Then the measurement vector, in the absence of measurement noise is given by

$$y = Ax. \quad (5)$$

Each measurement $y_i$ in the vector $y$ is the light intensity measured by the sensor, in the absence of measurement noise, when the transmittance of the aperture elements are programmed according to $a_{ij}, j = 1,...,N$, that is, the transmittance of aperture element $j$ is equal to $a_{ij}$.

*B. Measurement noise*

The measurement vector $y$ in Eq (5) is subject to measurement noise. We denote the vector of acquired values from the sensor in presence of measurement noise by $z$ and will establish a relationship between $z$ and $y$ by making some assumptions on the noise. We assume that two types of noise are present in the measurement vector $y$.

***Shot Noise***:

The shot noise is caused by statistical quantum fluctuations in the number of photons collected by the sensor, and it is modeled by the Poisson distribution. With the shot noise, the actual acquired value from the sensor for measurement $y_i$ is a random variable $\hat{y}_i$ given by

$$\hat{y}_i \sim P(y_i), \quad (6)$$

where $P(y_i)$ denotes the Poisson distribution with the mean $y_i$. We further assume that $\hat{y}_i, i = 1,...,N$ are independent random variables. By the definition of Poisson distribution, we have

$$E(\hat{y}_i) = \text{var}(\hat{y}_i) = y_i, i = 1,...,N. \quad (7)$$

In Eq (7), $E(\cdot)$ and $\text{var}(\cdot)$ denote the expected value and the variance, respectively.

***Additive Noise***:

This type of noise has a fixed variance independent of light intensity. This noise includes thermal noise and quantization noise in the measurements, and is modeled by a random variable $\varepsilon_i$ for measurement $y_i$. We further assume that $\varepsilon_i, i = 1,...,N$ are independent and identically distributed random variables with zero mean and a variance of $\sigma^2$, i.e.,

$$E(\varepsilon_i) = 0, \text{var}(\varepsilon_i) = \sigma^2, i = 1,...,N. \quad (8)$$

We only need to make assumptions in Eq (8), because the distribution of the additive noise itself is not important in our analysis.

Therefore, under the assumption of the above two types of noises, the actual acquired value for $y_i$, read from the sensor in the presence of the measurement noise, can be written as

$$z_i = \hat{y}_i + \varepsilon_i, i = 1,...,N, \text{ or } z = \hat{y} + \varepsilon. \quad (9)$$

The vector $z$ of Eq (9) is the actual acquired data of the measurement vector $y$ from the sensor in the presence of measurement noise.

Although it can easily shown that the SNR of the measurement vector $z$ is quite high, much higher than the SNR of the pinhole aperture imaging, we are not interested in the SNR of the measurement vector $z$ itself, because the measurements are only intermediate values, and it is in the SNR of the final image reconstructed from the measurements that we are interested.

*C. Reconstructed image*

The virtual image of a scene is defined by $x$ in equations (1) and (2). In LCI, $x$ is not acquired directly, so that it must be reconstructed from the acquired measurement vector $z$. Reconstruction algorithms are well known in compressive sensing literature, see for example [1] [13], but in the context of this paper, since our sensing matrix is a square, invertible matrix, the reconstruction can be performed simply by solving Eq (5) for $x$, with $y$ replaced by $z$.

Let $\tilde{x}$ be the image reconstructed from the acquired measurement vector $z$, i.e.,

$$\tilde{x} = A^{-1} z . \quad (10)$$

Then the goal is to find the SNR of the reconstructed image $\tilde{x}$.

*D. Signal to noise ratio*

First we define the SNR for a pixel value $\tilde{x}_i$. The signal power at the pixel is $x_i$ and the power of the measurement noise at the pixel is $\sqrt{\mathrm{var}(\tilde{x}_i)}$. Therefore the SNR at the pixel is defined to be

$$\mathrm{SNR}_{MNi}^{LCI} = \frac{x_i}{\sqrt{\mathrm{var}(\tilde{x}_i)}} . \quad (11)$$

Next, we consider the total signal power and total noise power in the entire image $\tilde{x}$. The total signal power is the integration of all light rays from the scene to the sensor when all aperture elements are opened (having transmittance of 1), referred to Figure 1, and it is therefore given by

$$X^0 = \sum_{i=1}^{N} x_i . \quad (12)$$

The value $X^0$ defined in Eq (12) can be considered to be the brightness of the scene as seen by the sensor, and it is only a function of the lighting of the scene and the field of view of the LCI architecture, in particular, it is independent of the number of pixels in the image.

The total SNR of LCI due to measurement noise is defined as

$$\mathrm{SNR}_{MN}^{LCI} = \frac{X^0}{\sqrt{\sum_{i=1}^{N} \mathrm{var}(\tilde{x}_i)}} . \quad (13)$$

In Eq (13), the numerator is the total signal power in the image and the denominator is total noise power in the image.

The following is the main result of this paper, and its proof is given in Appendix.

*Proposition 1.*

*If the sensing matrix is the modified Hadamard matrix given in Eq* (4), *then the following expressions hold for the SNR of the reconstructed image $\tilde{x}$:*

$$\mathrm{SNR}_{MNi}^{LCI} = \frac{\sqrt{N} x_i}{\sqrt{\left(2 - \frac{4}{N}\right) X^0 + \left(4 - \frac{4}{N}\right)\sigma^2}}$$

$$\geq \frac{\sqrt{N} x_i}{\sqrt{2 X^0 + 4\sigma^2}} \quad (14)$$

$$\mathrm{SNR}_{MN}^{LCI} = \frac{X^0}{\sqrt{\left(2 - \frac{4}{N}\right) X^0 + \left(4 - \frac{4}{N}\right)\sigma^2}}$$

$$\geq \frac{X^0}{\sqrt{2 X^0 + 4\sigma^2}} \quad (15)$$

*where $\sigma^2$ is the variance of the additive noise given in Eq (8).*

The lower bound for the total SNR in Eq (15) is only a function of the brightness $X^0$ and the power of the additive noise, $\sigma^2$. In the denominator, the value $\sqrt{X^0}$ represents the total power of shot noise, and $\sigma$ represents the power of additive noise when the sensor acquires each measurement.

An important observation from Proposition 1 is that the lower bound in Eq (15) is independent of the image resolution, i.e., the number of pixels, $N$. The total SNR of LCI due to measurement noise is bounded below by a constant with respect to the image resolution. In particular, it does not reduce when the image resolution $N$ increases.

On the other hand, even though the factor $\sqrt{N}$ appears in the numerator of (14), it does not necessarily imply that $\mathrm{SNR}_{MNi}^{LCI}$ increases with the resolution of the image. This is because for a given scene, even though the brightness $X^0$ is independent of image resolution, the pixel value $x_i$ itself may be a function of the image resolution. Therefore, the numerator of (14), $\sqrt{N} x_i$, may increase, remain constant or decrease as $N$ increases. However, if a pixel $x_i$ represents a point source, such as a distant star, it is independent of the resolution, and the SNR at that pixel increases with the resolution.

IV. COMPARISON

In this section, we present the SNR results for two other imaging architectures: digital camera with the pinhole aperture imaging (PAI), and the digital camera with lens aperture imaging (LAI).

We assume that images in all architectures have the same resolution, i.e., the same number of pixels, $N$, which means that the number of aperture elements in LCI is the same as the number of sensors in PAI and LAI. We further assume that the scene and the field of view of the images are the same in all architectures to be compared, and the field of view does not change with the resolution $N$. A corollary of these assumptions is that the brightness of the scene, $X^0$ defined in Eq (12), is independent of resolution $N$.

We will compare the SNR of LCI due to measurement noise with each of the PAI and LAI and show that the former outperforms both the PAI and LAI if the image resolution is high enough, i.e., if $N$ is large enough.

*A. Measurement noise*

Similar to the previous section, two types of noises are modeled in PAI and LAI. What is different in this section is that here, the image is acquired as pixels by an array of sensors. For pixel $i$, the acquired pixel value by the corresponding sensor is $\tilde{x}_i$, which is different from the true pixel value $x_i$ due to the measurement noise. Using the same treatment as Eq (9), the acquired, noisy pixel value $\tilde{x}_i$ is given by

$$\tilde{x}_i = \hat{x}_i + \delta_i, i = 1,...,N, \text{ or } \tilde{x} = \hat{x} + \delta, \quad (16)$$

where $\hat{x}_i, i = 1,...,N$ are independent random variables with Poisson distribution, with the property

$$E(\hat{x}_i) = \text{var}(\hat{x}_i) = x_i, i = 1,...,N, \quad (17)$$

and $\delta_i, i = 1,...,N$ are independent and identically distributed random variables, with

$$E(\delta_i) = 0, \text{var}(\delta_i) = \rho^2, i = 1,...,N. \quad (18)$$

In Eq (18), we allow the additive noise to have a different power than that in (8) because the sensors may have different operating dynamic ranges in different architectures.

In the following, we will find the SNR in the acquired image $\tilde{x}$.

*B. Comparison with pinhole aperture imaging*

The LCI of Figure 1 is closely related to the PAI as illustrated in Figure 3.

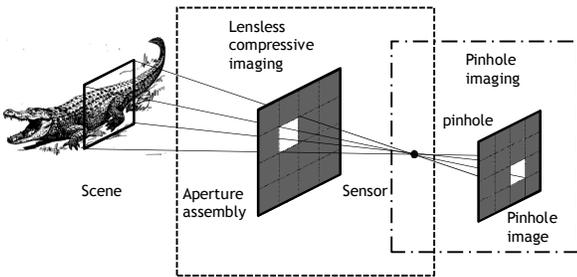

**Figure 3.** Relationship between lensless compressive imaging (LCI) and pinhole aperture imaging (PAI).

Let us consider the special case where the sensing matrix used in the LCI is the identity matrix. In other words, each measurement from the sensor of Figure 1 is made when only one of the aperture elements is open and all others are closed. Then a measurement is equivalent to a pixel value in the PAI when the pinhole is placed at the location of sensor, see Figure 3. In this case, the noise characteristics in LCI and PAI are exactly the same. That is, if the sensor in the LCI is the same as the sensors in the PAI, and if the identity matrix is used as the sensing matrix in LCI, then the two architectures have the same SNR. However, the result of Proposition 1 is obtained because the modified Hadamard matrix is used as sensing matrix instead of the identity matrix.

For the pinhole aperture imaging, we have the following result on the SNR of the image due to measurement noise. The proof is given in Appendix.

***Proposition 2.***

*The SNR of the image $\tilde{x}$ in pinhole aperture imaging is given by*

$$\text{SNR}_{MNi}^{PAI} = \frac{x_i}{\sqrt{x_i + \rho^2}}, \quad (19)$$

$$\text{SNR}_{MN}^{PAI} = \frac{X^0}{\sqrt{X^0 + N\rho^2}} \quad (20)$$

*where $X^0$ is the total signal power given in Eq (12), and $\rho^2$ is the variance of the additive noise given in Eq (18).*

*Furthermore, the following estimate holds for the ratio of $\text{SNR}_{MN}^{LCI}$ over $\text{SNR}_{MN}^{PAI}$*

$$\frac{\text{SNR}_{MN}^{LCI}}{\text{SNR}_{MN}^{PAI}} \geq \frac{\sqrt{X^0 + N\rho^2}}{\sqrt{2X^0 + 4\sigma^2}} \approx \frac{1}{\sqrt{2}}\sqrt{1 + \left(\frac{\sqrt{N}\rho}{\sqrt{X^0}}\right)^2}. \quad (21)$$

An important observation from Proposition 2 is that the total SNR in PAI, $\text{SNR}_{MN}^{PAI}$, is not only a function of $X^0$ and $\rho^2$, like Proposition 1, but also a function of the image resolution $N$, unlike Proposition 1. The significance of the Proposition 2 is that the SNR of PAI decreases as the image resolution increases.

It is more revealing if we consider the ratio of the SNRs for LCI and PAI, as given in Eq (21), which shows that the SNR of the LCI is higher than that of the PAI by an order of $\sqrt{N}$. One corollary is that no matter what sensors and quantization levels are used in two architectures (which determine the relative sizes of $\sigma$ and $\rho$), the LCI will always outperform the PAI if the image resolution is high enough, i.e., if $N$ is large enough.

Although Eq (21) shows that $\text{SNR}_{MN}^{LCI}$ can be arbitrarily better than $\text{SNR}_{MN}^{PAI}$ as $N$ increases, in reality, there is a practical limit on the size of $N$. Therefore, in addition to the discussion of the asymptotical behavior of Eq (21) as $N$ approaches to infinity, we also consider the general case of any given $N$. In the numerator in the last fraction of Eq (21), $\sqrt{N}\rho$ is the power of the total additive noise in the PAI, and in the denominator, $\sqrt{X^0}$ is the power of the total shot noise in PAI. Eq (21) shows that if the total additive noise in the PAI image is higher than the total shot noise, then LCI has a better SNR than PAI. The SNRs of the two architectures are the same if the total shot noise is equal to the total additive noise in the PAI. Therefore, whether LCI or PAI has higher SNR depends on whether the total additive noise or the total shot noise is higher in PAI. This is shown in Figure 4.

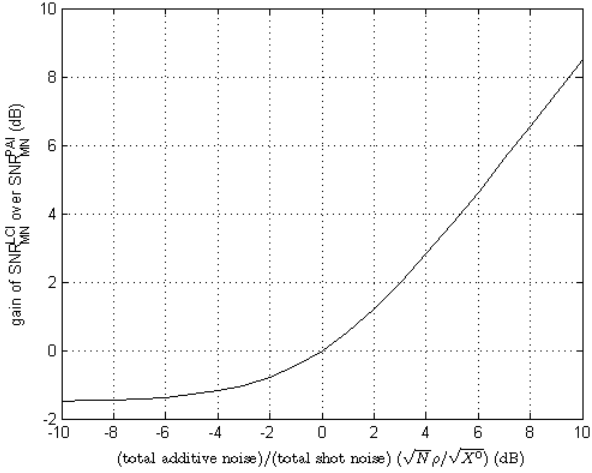

**Figure 4.** $\text{SNR}_{MN}^{LCI}/\text{SNR}_{MN}^{PAI}$ as a function of $\sqrt{N}\rho/\sqrt{X^0}$, the total additive noise over the total shot noise in the PAI image.

A few remarks are in order. First, note that while the total shot noise is a constant, the total additive noise in PAI is an increasing function of $N$, and therefore, an increase the image resolution increases performs of LCI as compared to PAI.

Secondly, as shown in Eq (21) and Figure 4, in the worst case scenario, the SNR of LCI can only be lower than that of PAI by a factor of $\sqrt{2}$, which is about 1.5 dB. That is, the total SNR of LCI can be no more than 1.5dB worse than that of PAI under any circumstance.

Third, the SNR of LCI is much better than PAI if the total shot noise is low, which happens if the scene is faint. This shows that LCI can have much better performance in low lighting environment, such as in surveillance or astronomy. In other words, LCI outperforms PAI in cases where SNR is concerned most, which is when the shot noise is low. When the shot noise is high, the SNR is high also, causing very little concern about it.

*C. Pixel signal to noise ratio*

We now consider pixel level SNR and compare the pixel SNRs of LCI and PAI, $\text{SNR}_{MNi}^{LCI}$ and $\text{SNR}_{MNi}^{PAI}$ given by Eq (14) and Eq (19), respectively.

First, in LCI, the noise in each pixel, the denominator in (14), is independent of the pixel value $x_i$, i.e., all pixels have the same amount of noise in the reconstructed image, even though the shot noise is included in the calculation. In other words, the total noise in the reconstructed image is evenly spread out among all pixels. This feature is useful in an application where the scene is dim, and only the details of a few small, relatively brighter, regions are of interest, such as imaging of distant stars in the sky, because the higher shot noise in a brighter region of interest (ROI) will be distributed to other darker areas of no interest, resulting a better SNR in the ROI.

Next, let us examine what happens at the pixels if the total SNR of LCI is less than that of PAI. Since the total SNR of LCI becomes the lowest if the additive noise in PAI is zero, i.e., when only shot noise is present, we compare the pixel SNRs under the assumption that there is no additive noise, i.e.,
$$\rho = \sigma = 0 . \qquad (22)$$

Under this assumption, the following equation can be easily derived from Propositions 1 and 2.
$$\frac{\text{SNR}_{MNi}^{LCI}}{\text{SNR}_{MNi}^{PAI}} = \sqrt{\frac{x_i}{2X^0/N}} . \qquad (23)$$

The numerator in (23) is the pixel value, and the denominator is twice of the average pixel value which is $X^0/N$. This shows that even if the total SNR of LCI is lower, a pixel in the LCI image can still have higher SNR if the pixel value is higher than twice of the average pixel value, i.e.,
$$\text{SNR}_{MNi}^{LCI} \geq \text{SNR}_{MNi}^{PAI}, \text{ if } x_i \geq 2X^0/N . \qquad (24)$$

Eq (24) confirms the earlier assertion that the brighter pixels in the LCI image have higher SNR that those in the PAI.

*D. Comparison with lens aperture imaging*

To compare the LCI with the LAI, we need to overcome one technical hurdle that so far, we have been assuming the sensor in LCI of Figure 1 has an infinitesimal size, while a lens has a finite, nonzero, size. In order to do the comparison, we need to assume that the sensor in the LCI has a finite, nonzero, size too.

A nonzero size sensor in LCI will introduce blurring into the image, but the blurring can be removed or reduced during reconstruction [14]. Nevertheless, it is out of scope of this paper to consider the blurring or how to reduce the blurring. We will simply consider the blurred image as the desired image that we want to acquire, and there is no loss of rigor in doing so. This is because in the LAI, a perfectly non-blurred image is obtained only when the image plane, i.e., the plane of sensors, is placed exactly at the focal plane of the lens. In reality, this would never be possible, because just like we can never make an infinitesimal sensor in realty, we can never place an image plane at the exactly location of the focal plane in reality. Therefore, in LAI there is always a blurring in the image due to the imaging plane not exactly being at the focal plane, even if we assume that the lens itself is perfectly made, which is also never be possible in realty.

Therefore, in this subsection, when comparing with LAI, we assume the sensor in LCI has a nonzero size, and we compare it with an LAI in which the image plane is not placed at the focal plane so that the images in both architectures have exactly the same amount of blurring. This is illustrated in Figure 5.

As shown in Figure 5 (a), we assume that the sensor in LCI has a nonzero size, and its area is given by $S_{sensor}$. The area of the lens in LAI is given by $S_{lens}$ as shown in Figure 5 (b). The areas $S_{sensor}$ and $S_{lens}$ may be very different, but for the two architectures to have the same amount of blurring, the point spread functions, as illustrated in Figure 5, are assumed to be the same. In other words, we assume the imaging plane of LAI is placed appropriately away from the focal plane so that the point spread function matches that of LCI due to non-zero size sensor.

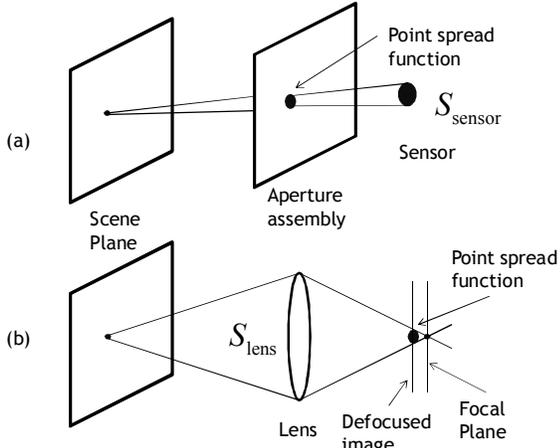

**Figure 5.** Blurred images in LCI and LAI. (a) Blurring due to nonzero size sensor in LCI. (b) Blurring due to displacement of image plane away from the focal plane in LAI.

When comparing with PAI, we also assume that the size of the pinhole is nonzero, and the area of the hole is same as the area of the sensor in LCI, which is $S_{sensor}$. Note that $S_{sensor}$ is area of the sensor in LCI, or the area of pinhole in PAI, it is not the area of the sensors in PAI or LAI.

We define the gain of lens as
$$g = S_{lens}/S_{sensor} \ . \quad (25)$$
In comparing the images in LAI and PAI, we find that they are the same, with the exception that the scene appears $g$ times brighter in LAI due to the gain of lens. This is because the amount of light rays arriving at the image plane when a lens is used is $g$ times more than that when a pinhole is used since the area of the lens is $g$ times larger than the pinhole. Consequently, a scene which is found to have the brightness of $X^0$ in PAI will be found to have the brightness of $gX^0$ in LAI. Therefore, the following Proposition follows directly from Proposition 2 by replacing $X^0$ by $gX^0$, and $x_i$ by $gx_i$.

*Proposition 3.*

The SNR of the image $\tilde{x}$ in lens aperture imaging is given by
$$\text{SNR}_{MNi}^{LAI} = \frac{gx_i}{\sqrt{gx_i + \rho^2}}, \quad \text{SNR}_{MN}^{LAI} = \frac{gX^0}{\sqrt{gX^0 + N\rho^2}} \ . \quad (26)$$

Further, the following estimate holds for the ratio of $\text{SNR}_{MN}^{LCI}$ over $\text{SNR}_{MN}^{LAI}$
$$\frac{\text{SNR}_{MN}^{LCI}}{\text{SNR}_{MN}^{LAI}} \geq \frac{\sqrt{gX^0 + N\rho^2}}{g\sqrt{2X^0 + 4\sigma^2}} \approx \frac{1}{\sqrt{2g}}\sqrt{1 + \left(\frac{\sqrt{N}\rho}{\sqrt{gX^0}}\right)^2} \ . \quad (27)$$

Comparing Propositions 2 and 3, we find that the SNR in LAI is higher than that in PAI because of the gain of the lens, $g$. Despite having a higher value, the SNR of LAI exhibits a same characteristic as that of PAI, namely, the SNR decreases as the image resolution $N$ increases.

Equation (27) shows that for given configurations of the two architectures, the LCI can have a higher SNR than the LAI if the image resolution is high enough, or if the scene is dim enough. Specifically, LCI has a higher SNR if the total additive noise $\sqrt{N}\rho$ in LAI, which is an increasing function of the image resolution $N$, is higher than the total shot noise $\sqrt{gX^0}$ in LAI, which is a constant independent of image resolution.

## V. DISCUSSION

In this section, we provide discussions and explanations to shed more insight into the understanding of the LCI.

*A. SNR and image resolution*

The key result from the analysis of the previous sections is that the total SNR in LCI is not a function of the image resolution, or the number of pixels in the image, while that in PAI or LAI is. To better understand the reason behind this, let us first review the noise behavior in PAI.

In PAI, the number of sensors in the image plane determines the image resolution, one sensor for each pixel. Each sensor introduces an additive noise of power $\rho$ into the image, and since the noise from the sensors are independent random variables, the total power of noise in the image is therefore $\sqrt{N}\rho$ with $N$ sensors. Same is true for LAI.

Lensless compressive imaging works differently. While the sensor in LCI also introduces an additive noise of power $\sigma$ when making each measurement, the measurements themselves are not uncorrelated. In the reconstruction process, the additive noise from all measurements contributes to a total noise of power $\sqrt{N}\sigma$, similar to PAI or LAI, but the signal itself, being present in each measurement, also adds up coherently if an appropriate sensing matrix, such as the modified Hadamard matrix, is used, resulting in a gain of $\sqrt{N}$ in the signal power. The gain in the signal cancels out the gain in the additive noise, and hence, the total noise is independent of the image resolution. This advantage only comes with a correctly chosen sense matrix. For example, there is no signal gain if the identity matrix is used as the sensing matrix, which comes with no surprise because the SNR in this case is equivalent to the PAI, as previously noted.

*B. Sensor-time constant*

There is also a justification from physics on why LCI can have a higher SNR. To acquire the image in LCI, the sensor needs to make a series of snapshots, each taking a unit amount of time $\Delta t$ as defined in Eq (1). For an image of resolution $N$, the total time duration it takes to acquire all independent measurements in LCI is $N\Delta t$. On the other hand, the time duration for acquiring an image in LAI or PAI is only $\Delta t$.

Therefore, the SNR gain of LCI over PAI or LAI is realized by taking the advantage of longer exposure time. But does the longer exposure time create an unfair advantage for LCI over PAI or LAI? The answer is no. Even though LCI takes longer time, it uses fewer sensors: only one sensor in LCI as opposed to $N$ sensors in LAI or PAI. Although the exposure time is $N$ times longer, the number of sensors is $N$ times smaller in LCI than in LAI or PAI. Therefore, all three architectures share the same constant which is the number of sensors multiplied the exposure time, i.e., the following holds in all three architectures:

$$(\text{number of sensors}) \times (\text{exposure time}) = N\Delta t. \quad (28)$$

Because of the sensor-time constant, it would be misleading to make a general statement about which architecture is preferable simply on the basis of the amount of exposure time; whether a shorter exposure time or fewer sensors is more desired depends on the application. For example, the LCI is more advantageous in applications where sensors are very expensive, but the scene does not change rapidly, such as in astronomy.

Furthermore, in LAI or PAI, even though the exposure time is $\Delta t$, the pixel values must be read out from the $N$ sensors sequentially, and the overhead for reading out the large number of the pixel values sequentially may consume a substantial amount of time too.

*C. Relation with coded aperture*

The coded aperture architecture has a higher SNR than the PAI by coherently adding signals resulted from the coded pinhole apertures. However, the coded aperture architecture is a special case of LCI with multiple sensors [11]. When multiple sensors are used in LCI, and when only one measurement from each sensor is used in the reconstruction, the resulting image is the same as that from the coded aperture. We refer to [11] for more details on the reconstruction of an image using measurements from multiple sensors.

## VI. SIMULATION

We present some simulation results in this section.

*A. Total SNR as a function of image resolution*

In the first simulation, we demonstrate the behavior of the total SNRs of LCI and PAI, $\text{SNR}_{MN}^{LCI}$ and $\text{SNR}_{MN}^{PAI}$, as a function of image resolution $N$.

First, we assume a scene has a fixed brightness $X^0$, in terms of number of photons. Then these photons are randomly assigned to the $N$ pixels of the image $x$ with a uniform distribution, so that the total number of photons in the image is $X^0$. The Poisson distribution is used to create shot noise, and the Gaussian distribution of variance $\sigma^2 = \rho^2$ is used to create additive noise. In LCI, the noise is added to the measurements, and the reconstructed image $\tilde{x}$ is obtained from the contaminated measurements by inverting the sensing matrix. In PAI, the noise is added to the pixels of $x$ to obtain the contaminated image $\tilde{x}$. We then compute the total SNR of the entire image $\tilde{x}$ for different values of image resolution $N$, and plot the results, together with the values obtained from theoretical analysis of previous sections. The results are presented in Figure 6.

In the simulation of Figure 6, the following parameters are used

$$X^0 = 10^7, \sigma = \rho = 5. \quad (29)$$

It can be observed that the SNR in LCI is a constant with respect to the image resolution $N$, while SNR in PAI decreases as the image resolution increases. Furthermore, the simulation results match very well with the theoretical analysis of the previous sections.

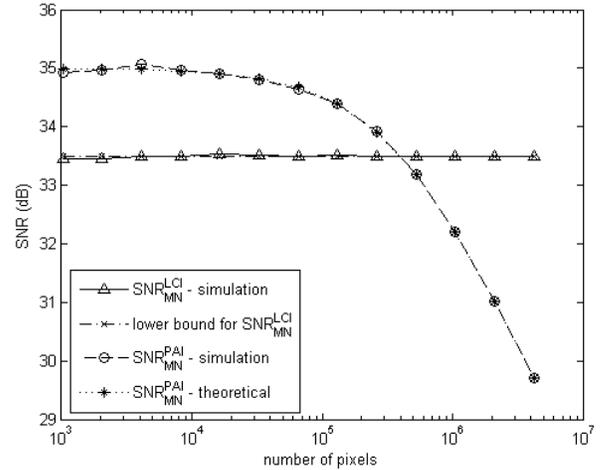

**Figure 6.** SNR of LCI and PAI as functions of image resolution (number of pixels). The lower bound for $\text{SNR}_{MN}^{LCI}$ is given by (15) and the theoretical result for $\text{SNR}_{MN}^{PAI}$ is given by (20).

*B. Shot noise only*

Since the worst performance of LCI as compared to PAI is when the additive noise is absent, i.e., there is only shot noise, we next perform simulations under the assumption that no additive noise is present, i.e.,

$$\sigma = \rho = 0. \quad (30)$$

The simulation is performed for five images: Earth, Sun, Parking Lot, Predator and Highway. These images were downloaded from publicly available website pages, found through Google image search.

For first two images, Earth and Sun, the pixel values of the original images are modified to yield a given amount of average photons per pixels. The Poisson distribution is used to introduce shot noise to the measurements for LCI and to the pixels for PAI.

For each of the two images, we define a region of interest (ROI). We will show the image of ROI and present the SNRs in the ROI for LCI and PAI, $\text{SNR}_{ROI}^{LCI}$, $\text{SNR}_{ROI}^{PAI}$, respectively.

**Earth**

The simulation result for Earth is shown in Figure 7. The original image is shown on top, and the bottom shows the ROI for the reconstructed image in LCI, the original image and the noisy image in PAI, respectively from left to right. The parameters used in, and the results from, the simulation are summarized in the following:

$$X^0/N = 0.2, \text{SNR}_{ROI}^{LCI} = 35.0 dB, \text{SNR}_{ROI}^{PAI} = 30.7 dB. \quad (31)$$

This simulation demonstrates that in the absence of the additive noise, which is the worst case for the SNR of LCI as compared to PAI, the SNR of LCI is more than 4dB better than that of PAI in the ROI.

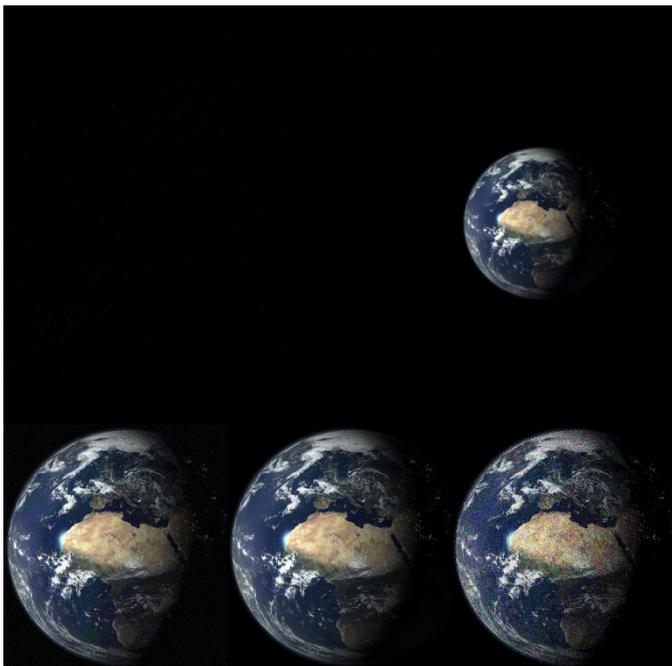

**Figure 7.** Earth. Top: the original image. Bottom: details of the region of interest (ROI). Bottom left: reconstructed image from LCI, bottom middle: original, bottom right: noisy image in PAI. Average number of photons: 0.2 photons/pixel. SNR of ROI in LCI=35.0dB (bottom left). SNR of ROI in PAI=30.7dB (bottom right)

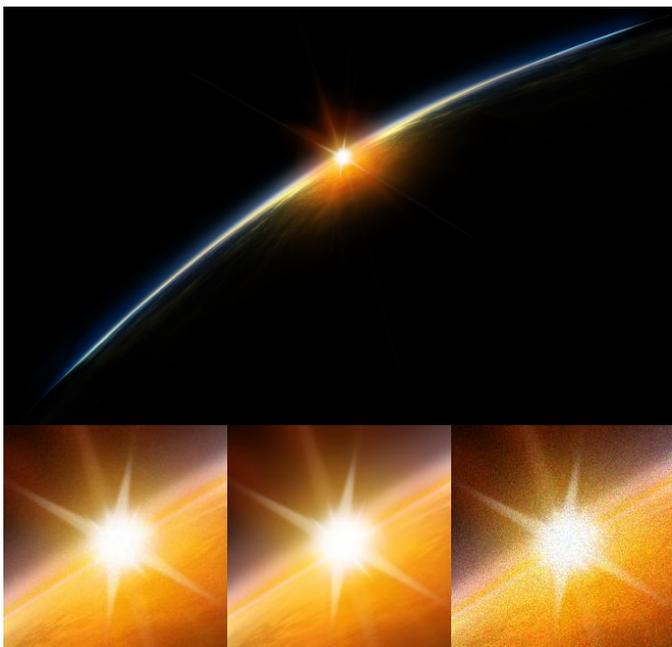

**Figure 8.** Sun. Top: the original image. Bottom: details of the region of interest (ROI). Bottom left: reconstructed image from LCI, bottom middle: original, bottom right: noisy image in PAI. Average number of photons: 5 photons/pixel. SNR of ROI for LCI=40.6dB (bottom left). SNR of ROI for PAI=35.7dB (bottom right)

**Sun**

The simulation result for Sun is shown in Figure 8. The original image is shown on top, and the bottom shows the ROI for the original image, and the reconstructed image in LCI and contaminated image in PAI. The following summarizes the assumptions and results

$$X^0/N = 5, \text{SNR}_{ROI}^{LCI} = 40.6dB, \text{SNR}_{ROI}^{PAI} = 35.7dB. \quad (32)$$

The results show that SNR of LCI is about 5dB better than that of PAI in the ROI.

In the next group of simulations, for each image, we examine each pixel in the image and compare its value with the value that is twice of the average pixel value of the image, i.e., $2X^0/N$. This provides a metric of how the pixel SNR of LCI compares with that of PAI given by Eqs (23) and (24).

For each pixel in an image, we compute the ratio of the pixel value, $x_i$, over the twice average pixel value, $2X^0/N$, and express it in dB. Then the dB values are treated as pixel values of an image to be displayed in grayscale.

The results for the images, Parking Lot, Predator and Highway are presented below.

**Parking Lot**

Figure 9 is the result for a night surveillance image of a parking lot. The result shows that in the ROI, the person's body, the SNR of LCI can be up to 6dB higher than that of PAI, even though no additive noise is present.

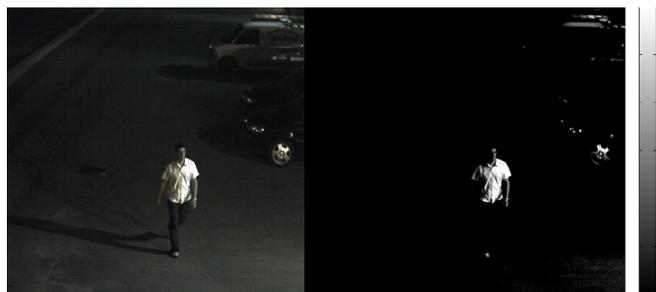

**Figure 9.** Parking Lot. Left: the original image. Middle: dB value of how much higher the pixel value is compared with $2X^0/N$, twice of average pixel value. Right: the grayscale of the dB values in the middle image. For example, the darkest pixel represents values of less than or equal to 0dB and the brightest pixel represents value of about 6dB.

**Predator**

Figure 10 shows the result for a night vision of a predator drone. The result shows that the pixels on the drone have up to 5 or 6dB higher SNR in LCI than in PAI.

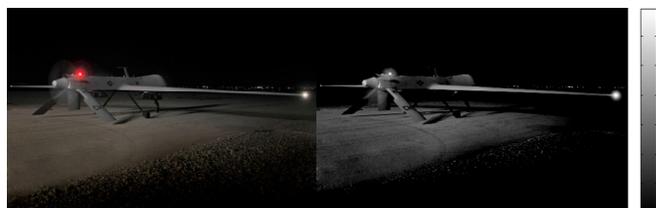

**Figure 10.** Predator. Left: the original image. Middle: dB value of how much higher the pixel value is compared with $2X^0/N$, twice of average pixel value. Right: the grayscale of the dB values in the middle image. For example, the darkest pixel represents values of less than or equal to 0dB and the brightest pixel represents value of about 7dB.

**Highway**

Figure 11 shows the result for a scene of highway accident. The result shows that in the most important areas, the road sign and the police car license plate, the pixels have up to 7dB higher SNR in LCI than in PAI.

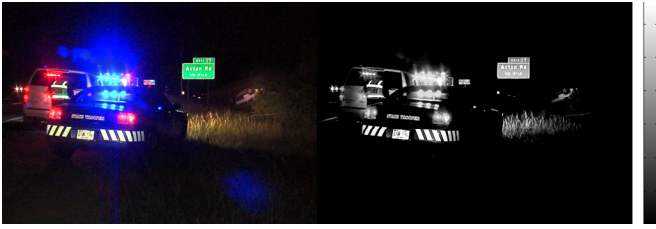

**Figure 11.** Highway. Left: the original image. Middle: dB value of how much higher the pixel value is compared with $2X^0/N$, twice of average pixel value. Right: the grayscale of the dB values in the middle image. For example, the darkest pixel represents values less than or equal to 0dB and the brightest pixel represents value of about 6.5dB.

**Percentage of pixels that have higher SNR in LCI**

Finally, we present an overview of the amount of pixels with different values of SNR in the images in LCI and PAI. The result is summarized in Figure 12.

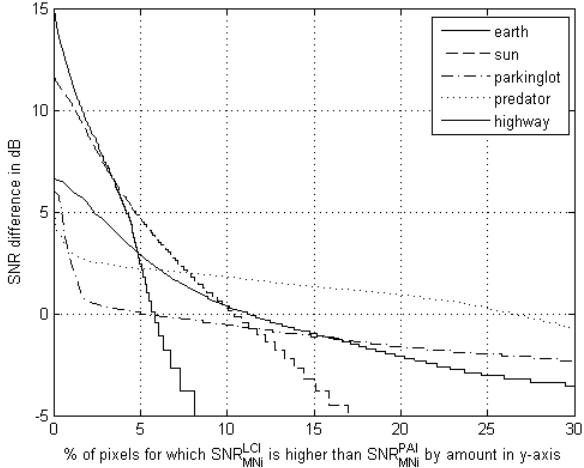

**Figure 12.** Percentage of pixels for which the pixel SNR of LCI is better than that of PAI by a given amount under the assumption that there is no additive noise.

The result in Figure 12 is obtained by computing the ratio, in dB, that is given in Eq (23) at each pixel. The ratio of (23) is the metric of comparison between the pixel SNR of LCI and PAI. After the ratios are computed for all pixels of an image, they are sorted in descending order, and plotted against the position index normalized by the total number of pixels $N$. One curve is presented for each of the five images.

For each point on a curve, the y-coordinate is the dB by which the pixel SNR is higher in LCI than in PAI. A negative dB means the SNR in PAI is higher than in LCI. The x-coordinate represents the percentage of pixels in the image whose SNR in LCI is higher than that of PAI by that dB amount. For example, in the image of Predator, shown by the dotted line in Figure 12, there are more than 25% pixels that have higher SNR in LCI than in PAI (the curve is above 0dB for all x-values below 25), while in Earth image, shown by a solid curve, there are only about 5% of pixels have higher SNR in LCI, but some pixels have close to 15dB better SNR in LCI than in PAI.

## VII. CONCLUSION

We performed SNR analysis for the lensless compressive imaging (LCI) under presence of the measurement noise, and compare it with that of pinhole aperture imaging (PAI) and lens aperture imaging (LAI). The main result is that the SNR in the LCI is independent of the resolution of the image, while that for either PAI or the LAI decreases as the number of pixels increases. Consequently, LCI can have a better SNR performance than either PAI or LAI when the image resolution is high enough.

For any given image resolution, whether the LCI or PAI has a better SNR depends on whether the total additive noise or shot noise dominates. The LCI has a better SNR than PAI if the additive noise is higher than shot noise, which is of more interest, because SNR is of most concern when the shot noise is low due to the fact that low shot noise also means low SNR.

## APPENDIX

*A. Proof of Proposition 1*

Let the sensing matrix $A$ be defined as the modified Hadamard matrix given in (4). Then the inverse of $A$ is given by

$$A^{-1} = \frac{2}{N}H - O, \quad O = [o_{ij}], \quad o_{ij} = \begin{cases} 1, & i = j = 1 \\ 0, & \text{otherwise} \end{cases} \quad (33)$$

In above $H$ is the Hadamard matrix.

We now proceed to compute the variance of $\tilde{x}$ given in (10). From (10) and (9), we have

$$x = A^{-1}z = A^{-1}\hat{y} + A^{-1}\varepsilon. \quad (34)$$

Since $\hat{y}$ and $\varepsilon$ are independent, we have

$$\text{var}(\tilde{x}) = \text{var}(A^{-1}\hat{y}) + \text{var}(A^{-1}\varepsilon). \quad (35)$$

The acquired measurement vector $z$ is given by (9). The Hadamard matrix $H$ has the property that the entries in the first row all have value 1, and therefore, the first measurement $z_1$ is the sum of all pixels in the object plane. $z_1$ has statistics different from other measurements $z_i$ which unnecessarily complicates the analysis. We therefore will remove the measurement $z_1$ from both the acquisition and the reconstruction. We are now left with $N-1$ equations, and to invert the matrix, we impose the condition that $x_1 = 0$, which can be accomplished, for example, by always blocking the first aperture element in the aperture assembly, or by not assigning the transmittance $a_{i1}$ (the first column of $A$) to any aperture element. This arrangement leads to the following conditions on the values of the first component of each of the vectors:

$$\text{var}(\hat{y}_1) = \text{var}(\varepsilon_1) = x_1 = 0. \quad (36)$$

Alternatively, the effect of (36) can be achieved by using a different sensing matrix $A_R$, of dimension $(N-1) \times (N-1)$, which is obtained from $A$ by removing the first row and the first column[1].

By using the expression of $A^{-1}$ from (33), and the assumption (36), the first term in (35) can be calculated as

$$\operatorname{var}(A^{-1}\hat{y}) = \operatorname{var}\left(\frac{2}{N}H\hat{y} - O\hat{y}\right) = \frac{4}{N^2}\operatorname{var}(H\hat{y})$$
$$= \left(\frac{4}{N^2}\sum_i \operatorname{var}(\hat{y}_i)\right)\vec{1} = \left(\frac{4}{N^2}\sum_{i \neq 1}\sum_j a_{ij}x_j\right)\vec{1} \quad (37)$$
$$= \left(\frac{4(N/2-1)}{N^2}\sum_j x_j\right)\vec{1} = \left(\frac{2N-4}{N^2}X^0\right)\vec{1}$$

In Eq (37), $\vec{1}$ denotes the vector of length $N$ with each of its component is equal to 1. Also in (37), we have used the property

$$\sum_{i \neq 1} a_{ij} = \frac{N}{2} - 1, j = 2, ..., N \quad (38)$$

Similarly, the second term in (35) is found to be

$$\operatorname{var}(A^{-1}\varepsilon) = \operatorname{var}\left(\frac{2}{N}H\varepsilon - O\varepsilon\right) = \frac{4}{N^2}\operatorname{var}(H\varepsilon)$$
$$= \left(\frac{4}{N^2}\sum_{i \neq 1}\operatorname{var}(\varepsilon_i)\right)\vec{1} = \left(\frac{4N-4}{N^2}\sigma^2\right)\vec{1} \quad (39)$$

From (37) and (39), we have

$$\operatorname{var}(\tilde{x}_i) = \frac{2N-4}{N^2}X^0 + \frac{4(N-1)}{N^2}\sigma^2, i = 1, ..., N, \quad (40)$$

which leads to

$$\sum_{i=1}^{N}\operatorname{var}(\tilde{x}_i) = N\left(\frac{2N-4}{N^2}X^0 + \frac{4(N-1)}{N^2}\sigma^2\right)$$
$$= \left(2 - \frac{4}{N}\right)X^0 + \left(4 - \frac{4}{N}\right)\sigma^2 \quad (41)$$

By using definition of $\operatorname{SNR}_{MN}^{LCI}$ from Eqs (13) and (41), we have

$$\operatorname{SNR}_{MN}^{LCI} = \frac{X^0}{\sqrt{\left(2 - \frac{4}{N}\right)X^0 + \left(4 - \frac{4}{N}\right)\sigma^2}}$$
$$\geq \frac{X^0}{\sqrt{2X^0 + 4\sigma^2}} \quad (42)$$

which completes the proof of Proposition 1.

### B. Proof of Proposition 2

The image with measurement noise in PAI is given by (16), from which we can find the variance of $\tilde{x}_i$ to be

$$\operatorname{var}(\tilde{x}_i) = \operatorname{var}(\hat{x}_i) + \operatorname{var}(\delta_i), \quad (43)$$

because $\hat{x}_i$ and $\delta_i$ independent random variables. From (17) and (18), we have

$$\operatorname{var}(\tilde{x}_i) = x_i + \rho^2, \quad (44)$$

which leads to

$$\operatorname{SNR}_{MN}^{PAI} = \frac{X^0}{\sqrt{\sum_i \operatorname{var}(\tilde{x}_i)}} = \frac{X^0}{\sqrt{X^0 + N\rho^2}}, \quad (45)$$

and the proof of Proposition 2.

---
[1] It can be shown that $A_R^{-1} = \frac{2}{n-2}\left(A_R - \frac{n-4}{n}(\Theta - A_R)\right)$, where $\Theta$ is the $(N-1) \times (N-1)$ matrix whose every element is equal to 1.

Further, in combining (42) and (45), we have

$$\frac{\operatorname{SNR}_{MN}^{LCI}}{\operatorname{SNR}_{MN}^{PAI}} \geq \frac{\sqrt{X^0 + N\rho^2}}{\sqrt{2X^0 + 4\sigma^2}} = \frac{\sqrt{1 + N\rho^2/X^0}}{\sqrt{2}\sqrt{1 + 2\sigma^2/X^0}}. \quad (46)$$

The last term in the denominator $2\sigma^2/X^0$ is very small because $X^0$ is the total brightness, or the square of power of the total shot noise in the entire image, while $\sigma^2$ is the power of the additive noise in one measurement. Eq (21) results from neglecting the last term in the denominator of (46).